\documentclass[letterpaper, 10 pt, conference]{ieeeconf}

\IEEEoverridecommandlockouts
\usepackage{amsmath,amsfonts,amssymb}
\usepackage{mathtools}
\usepackage{algorithm}
\usepackage{algorithmic}
\usepackage{graphicx}
\usepackage{booktabs}
\usepackage{multirow}
\usepackage{url}
\usepackage{cite}
\usepackage{bm}
\usepackage{xcolor}
\usepackage[font=small]{caption} %// added by varad, to make table caption small. used alot in ICRA/IROS.
\usepackage[hidelinks]{hyperref}

\makeatletter
\def\input@path{{./}{../}}
\makeatother
\usepackage{stfloats}

\title{\LARGE \bf
CARE: Communication-Free Planning using \\ Adaptive Regions of Engagement for Perimeter Defense}

\author{%
\parbox{\linewidth}{\centering
Prajwal Vijay*$^{1}$,
Vignesh Gurumurthy*$^{2}$,
Varad Vaidya$^{3}$,
Shubham Singhal$^{3}$,\\
Shridhar Velhal$^{4}$,
Kumaran Karthik$^{2}$,
Jishnu Keshavan$^{5}$,
Suresh Sundaram$^{2,3}$% 
}%
\thanks{Manuscript under review at ICRA 2027}%
\thanks{* indicates equal contribution}
\thanks{$^{1}$P. Vijay is with the Department of Electrical Engineering, Indian Institute of Technology, Madras, Chennai, India. P. Vijay is the corresponding author ({\tt\footnotesize ee23b057@smail.iitm.ac.in}).}%
\thanks{$^{2}$V. Gurumurthy, K. Karthik, S. Sundaram are with the Department of Aerospace Engineering, Indian Institute of Science, Bangalore, India.}%
\thanks{$^{3}$V. Vaidya, S. Singhal, S. Sundaram are with the Robert Bosch Center for Cyber Physical Systems, Indian Institute of Science, Bangalore, India.}%
\thanks{$^{4}$S. Velhal is with the Robotics and AI Group, Luleå University of
Technology, 97187 Luleå, Sweden.}%
\thanks{$^{5}$J. Keshavan is with the Department of Mechanical Engineering, Indian Institute of Science, Bangalore, India.}
}

\begin{document}

\maketitle
\thispagestyle{empty}
\pagestyle{empty}

%%%%%%%%%%%%%%%%%%%%%%%%%%%%%%%%%%%%%%%%%%%%%%%%%%%%%%%%%%%%%%%%%%%%%%%%%%%%%%%%
\begin{abstract}

In the perimeter defense problem, a team of defenders must intercept
intruders approaching a protected region before any of them crosses its
boundary. Decentralized solutions generally decide which defender engages which
intruder by communicating, through a consensus-based auction or a partition of
the perimeter that the team maintains jointly, so team performance inherits
the reliability of the communication channel. This paper asks whether the
assignment can instead be resolved from each defender's own observations. It
presents CARE (Communication-free planning using Adaptive Regions of
Engagement), a policy in which a defender senses over its full detection
radius but commits only within an engagement region whose radius is the
distance to its nearest observed teammate. Duplicate commitments are not
prevented but allowed to decay: the overlap that admits an intruder to the
candidate sets of two defenders shrinks as the two defenders converge on the
intruder. Each defender additionally maintains a wrapped-normal estimate of
the bearings at which it has observed arrivals, and drifts its home position
toward the density-weighted centroid of its own arc. This recovers the
benefit of an adaptive partition without requiring any communication.
Across 193 scenarios and 38,600 seed-paired episodes, CARE, without any
communication, marginally outperforms both baselines run with their
communication intact, by $0.012$ in capture rate over an adaptive
partitioning method and by $0.005$ over a decentralized bundle auction with
unrestricted communication. Removing the auction's messages lowers its
capture rate by $0.161$, whereas a single hop of target-intent sharing raises
CARE's by only $0.0004$, indicating that the coordination CARE forgoes is not
coordination it needs. CARE is demonstrated
on a Crazyflie team.

\end{abstract}

%%%%%%%%%%%%%%%%%%%%%%%%%%%%%%%%%%%%%%%%%%%%%%%%%%%%%%%%%%%%%%%%%%%%%%%%%%%%%%%%
\section{Introduction}
\label{sec:introduction}

Protecting a site from the air has become a problem of numbers. Small
unmanned aircraft are cheap, easy to launch and hard to stop, and a defense
that could once expect a single intruder must now expect many, arriving over
time and not all at once. The perimeter defense problem (PDP) captures this
setting: intruders converge on a protected region, and a smaller team of
defenders must intercept them at its boundary. Each defender has to decide
which intruder to engage and, when several are within reach, in what order.
Neither decision stays valid for long, since intruders keep arriving while
the episode runs and an interception feasible at one instant need not remain
so.

The PDP is classically posed as a differential game~\cite{isaacs1999differential}.
Solutions are available for a single engagement and for small teams, including
the case of defenders confined to the perimeter they
guard~\cite{von2020guarding}, a heterogeneous aerial--ground
pair~\cite{lee2020perimeter}, and cooperating defender
teams~\cite{shishika2020cooperative}. Because the joint state space grows rapidly with team size, the standard
approach is to decompose the multiplayer game into pairwise ones and to recast
the team problem as an assignment over their
outcomes~\cite{shishika2018local}. Reach-avoid analysis~\cite{fisac2015reach}
supplies those pairwise outcomes, and has been used to assign defenders to
intruders directly~\cite{chen2017multiplayer,yan2020task}. A complementary line
bounds what any policy can achieve against radially inbound Poisson arrivals,
for a single vehicle~\cite{bajaj2019dynamic} and for a
team~\cite{bajaj2024multivehicle}.
\cite{adler2024role} finds that the benefit of a heterogeneous defender team
depends on how much information its members hold: the information structure
is part of the problem. A survey is given in~\cite{shishika2020review}.

Recasting the PDP as an assignment problem is natural, because each intruder
defines a task with both a location and a deadline: reach the point at which
it will cross the perimeter, before it does. Under the spatio-temporal multi-task
assignment (STMTA) formulation~\cite{velhal2022decentralized,chopra2015spatio}
an episode becomes a sequence of such problems, and the formulation has since
been extended to dynamic resource allocation~\cite{velhal2022dynamic} and to
prioritized intruders~\cite{velhal2025priority}. Solving it in a decentralized manner is a
multi-robot task allocation
problem~\cite{korsah2013comprehensive,nunes2017taxonomy}, whose standard mechanism is a consensus-based auction in which
defenders construct bundles of tasks, bid on them, and exchange bids until the
allocation is conflict-free~\cite{choi2009consensus,brunet2008consensus}.
DMRST-MTA~\cite{velhal2022decentralized} applies this to the PDP and is the
first of our two baselines; we refer to it as \emph{the auction} below,
qualified by how much it is allowed to communicate. A different strategy dispenses with bidding altogether: adaptive
partitioning~\cite{macharet2020adaptive}, our second baseline, assigns each
defender an arc of the perimeter, estimates the density of arrival bearings,
and jointly slides the sector boundaries toward the estimated demand. Learning-based
assignment policies also exist for the
PDP~\cite{thousif2026decentralized,gurumurthy2026generalizable,lee2023graph}.
We do not compare against them since each resolves redundant assignments in a
consensus stage, and so presumes the communication this paper removes, and
each needs offline training on a scenario distribution.

The vulnerability common to these approaches is a dependence on communication
at the least convenient moment. An auction requires several rounds of messages
before its bundles settle, and it requires them while intruders are closing in. A
partition requires boundaries that remain consistent across the team; a
defender that loses contact with its neighbors is left planning against a
stale definition of its own sector. Adaptive partitioning makes the
dependence explicit by requiring each defender to share the state of its
virtual agent with its neighbors~\cite{macharet2020adaptive}. The dependence is measurable:
\cite{paulos2019decentralization} shows a learned team policy degrading as
its communication budget is tightened. Under jamming, range limits or
packet loss the failure mode is a common one: the defenders continue to fly,
but cease to agree on who is responsible for what. This fragility motivates
the question we address: \emph{can defenders construct effective interception
plans, and resolve the conflicts between them, without exchanging any
information?}

To address the question above, we propose CARE (Communication-free planning using Adaptive Regions
of Engagement),
a policy in which each defender acts on a self-contained observation
comprising its own state together with the teammates and intruders it detects,
and transmits nothing. Two mechanisms carry it. The first
separates sensing from commitment: a defender detects every intruder within
its detection radius but plans only against those inside a smaller
\emph{engagement region}, whose radius is the current distance to its nearest
detected teammate. Commitment therefore requires that an intruder be closer to
the defender than its neighbor is, a condition that two adjacent defenders
cannot both satisfy for long. Conflicts still arise, but they are dissipated
by the defenders' own motion, not by negotiation or auction
(Sec.~\ref{sec:method:conflicts}). The second mechanism relaxes the fixed
home positions that sustain coverage between engagements, since those are
well placed only under uniform arrivals. Each defender estimates the density
of bearings at which it has itself observed intruders appear, and drifts its
home toward the density-weighted centroid of its own arc.

CARE is evaluated on 193 scenarios spanning team size, arrival rate,
arrival concentration and defender-to-intruder speed ratio, at 200
seed-paired episodes each. Both baselines are run with their communication
intact: adaptive partitioning shares sector boundaries between neighbors,
and DMRST-MTA~\cite{velhal2022decentralized} exchanges bids across the whole
team. CARE, without any communication, marginally outperforms both, by
$0.012$ in capture rate over adaptive partitioning and by $0.005$ over the
auction. The margin over adaptive partitioning grows with team size and is negative
at three defenders. The most direct evidence for the central claim is an
ablation of CARE against itself: adding one hop of communication, in which
defenders share their current targets, raises its capture rate by $0.0004$,
whereas removing the auction's messages lowers its own by $0.161$.

The contributions are: (1)~a communication-free formulation of the
decentralized PDP in which selection, sequencing and motion all derive from a
single self-contained observation (Sec.~\ref{sec:problem}); (2)~an adaptive
engagement region that separates sensing from commitment, together with a
per-defender arrival-density estimator that moves home positions toward
observed demand (Sec.~\ref{sec:method}); (3)~an evaluation over 38{,}600
paired episodes, including the one-hop ablation (Sec.~\ref{sec:results}); and
(4)~a hardware demonstration (Sec.~\ref{sec:hardware}).

%%%%%%%%%%%%%%%%%%%%%%%%%%%%%%%%%%%%%%%%%%%%%%%%%%%%%%%%%%%%%%%%%%%%%%%%%%%%%%%%
\section{Problem Formulation}
\label{sec:problem}

\begin{figure}[t]
\centering
\includegraphics[width=0.88\columnwidth]{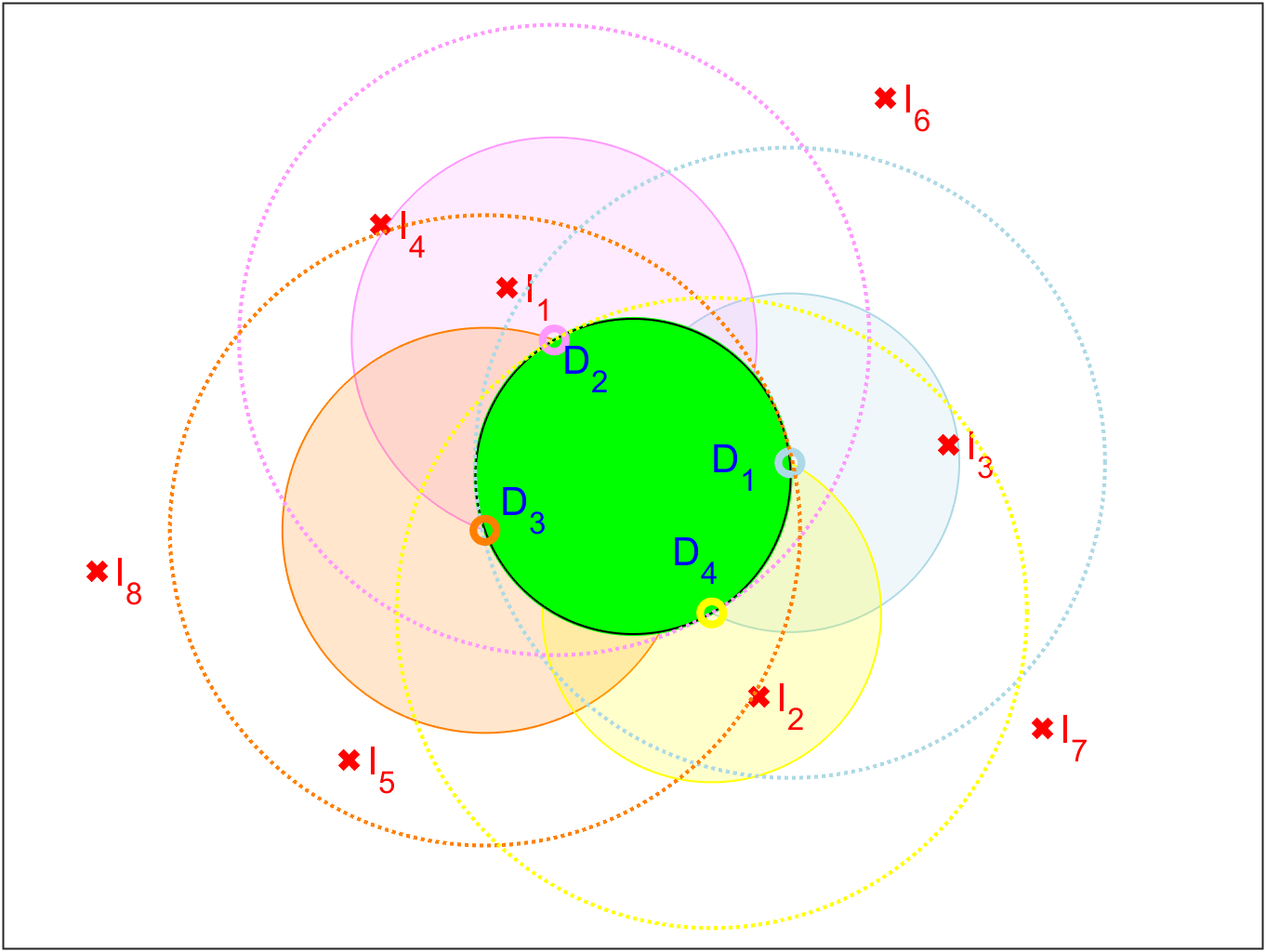}
\caption{The perimeter defense problem. Defenders $D_1,\dots,D_4$ (colored
circles) move on the perimeter of the protected territory; intruders
$I_1,\dots,I_8$ (\textcolor{red}{$\times$}) travel inward. Each defender
detects every intruder within $r_d$ (dotted circle in its color) but plans
only against those inside its engagement region (shaded disc), whose radius
$\rho_i(t)$ is the distance to its nearest detected teammate.}
\label{fig:setup}
\end{figure}

\subsection{Territory, defenders and intruders}

The protected territory is the disc
$\mathcal{T}=\{x\in\mathbb{R}^2:\lVert x\rVert\le R\}$ and the perimeter to be
defended is $\partial\mathcal{T}$. A team of $N_d$ defenders indexed by
$\mathcal{D}=\{1,\dots,N_d\}$ moves on that perimeter, so defender $i$ is
located by an angle $\phi_i(t)$,
\begin{equation}
p_i(t) = R\big(\cos\phi_i(t),\ \sin\phi_i(t)\big),
\qquad
R\,\lvert\dot\phi_i(t)\rvert \le v_D ,
\label{eq:defender_motion}
\end{equation}
with $v_D$ the maximum defender speed. Each defender carries a \emph{home
angle} $\theta_i(t)$, the position it returns to when it has no feasible
target. Defenders begin on their homes, $\phi_i(0)=\theta_i(0)$, uniformly
spaced around the perimeter. Whether those homes later move is a design
choice.

Intruders arrive during the episode rather than at its start. The $j$-th
intruder appears at time $t_j$ on a spawn circle of radius $R_s>R$ at bearing
$\psi_j$ and travels radially inward at speed $v_I$,
\begin{equation}
q_j(t) = \big(R_s - v_I (t-t_j)\big)\big(\cos\psi_j,\ \sin\psi_j\big),
\qquad t\ge t_j ,
\label{eq:intruder_motion}
\end{equation}
with arrival times following a Poisson process of rate $\lambda$ and bearings
drawn independently, either uniformly on $[-\pi,\pi)$ or from a von Mises
distribution $\mathcal{VM}(\mu,\kappa)$ when arrivals concentrate about a
preferred direction. Larger $\kappa$ concentrates the threat in a single
sector, the case in which uniformly spaced defenders are worst placed.
Equation~\eqref{eq:intruder_motion} is the nominal threat model, and is what
every method considered here uses to predict crossings; the benchmark
additionally includes zigzagging and spiralling intruders, for which that
prediction is wrong by a controlled amount.

Intruder $j$ is \emph{captured} if some defender comes within the capture
radius $r_c$ of it while it remains outside the territory, and \emph{breaches}
if it reaches the perimeter first. An episode runs for a fixed horizon $T$;
writing $N_c$ for the number captured and $N$ for the number that appeared, the
performance measure throughout is the capture rate
\begin{equation}
\eta = N_c/N \in [0,1],
\label{eq:capture_rate}
\end{equation}
and the objective is to maximize $\mathbb{E}[\eta]$. Fig.~\ref{fig:setup}
illustrates the setting.

We assume throughout that both teams are first-order and holonomic, with
bounded velocity as the control input, and that defenders are confined to $\partial\mathcal{T}$ and are homogeneous
in speed; this is the setting in both baselines as well. Intruders neither react to the defenders
nor cooperate with one another, each following a route fixed at spawn.
Detection within $r_d$ is exact and instantaneous, without misses, false
alarms or measurement noise, and a defender knows its own state exactly and
can associate an intruder across successive time steps. Capture is
instantaneous, removes the intruder and leaves the defender immediately
available; agents do not collide, and every defender executes the same policy
with the same parameters.

\subsection{Information structure}
\label{sec:problem:information}

The formulation until this point is standard. What distinguishes the present
work is what a defender is permitted to know at the moment it acts.

Defender $i$ has a single detection radius $r_d$ applying to teammates and
intruders alike, and at time $t$ observes
\begin{align}
\mathcal{V}_i(t) &= \{\, j : \lVert q_j(t)-p_i(t)\rVert \le r_d \,\},
\label{eq:visible_intruders}\\
\mathcal{K}_i(t) &= \{\, k \in \mathcal{D}\setminus\{i\} : \lVert p_k(t)-p_i(t)\rVert \le r_d \,\}.
\label{eq:visible_teammates}
\end{align}
Its \emph{self-contained observation} is the contents of those two sets
together with its own state,
\begin{equation}
o_i(t) = \Big(\,p_i(t),\ \{p_k(t)\}_{k\in\mathcal{K}_i(t)},\
\{(q_j(t),\dot q_j(t))\}_{j\in\mathcal{V}_i(t)} \Big),
\label{eq:observation}
\end{equation}
and a policy is \emph{communication-free} if the command of defender $i$
depends on its own observation history alone,
\begin{equation}
u_i(t) = \Pi\big(o_i(0{:}t)\big),
\label{eq:comm_free_policy}
\end{equation}
for a policy $\Pi$ common to all defenders. No bids, bundles, assignments, target identities or trajectories are
exchanged, and there is no central node. This is the constraint under which
CARE operates.

The two baselines occupy different points on the same axis. A consensus-based
method such as DMRST-MTA~\cite{velhal2022decentralized} closes the loop on
$u_i(t)=\Pi(o_i(0{:}t),\, m_{-i}(0{:}t))$, in which $m_{-i}$ denotes the
messages received from other defenders and several rounds of them are required
per planning step, whereas adaptive partitioning~\cite{macharet2020adaptive}
exchanges less traffic but still requires a partition upon which the team
agrees. We note that
\eqref{eq:comm_free_policy} prohibits a defender from being \emph{told} what
its neighbors intend, not from \emph{inferring} it; CARE relies on that
distinction throughout.

\subsection{Scale-free parameters}
\label{sec:problem:scalefree}

The dynamics \eqref{eq:defender_motion}--\eqref{eq:intruder_motion} contain no
intrinsic length or time scale: nothing depends on $R$ or $v_I$ separately,
only on the dimensionless groups
\begin{equation}
\underbrace{\frac{\lambda R}{v_I}}_{\text{load}},\quad
\underbrace{\frac{r_d}{R}}_{\text{sensing}},\quad
\underbrace{\frac{r_c}{R}}_{\text{capture}},\quad
\underbrace{\frac{v_D}{v_I}}_{\text{speed ratio}},\quad
\underbrace{\frac{T v_I}{R}}_{\text{horizon}} .
\label{eq:groups}
\end{equation}
The first of these counts arrivals per perimeter-crossing time and serves as
the difficulty parameter. Reporting against these groups, not against meters
and seconds, is what lets a policy characterized in simulation at $R=50$~m
fly at $R=0.5$~m without retuning (Sec.~\ref{sec:hardware}).

%%%%%%%%%%%%%%%%%%%%%%%%%%%%%%%%%%%%%%%%%%%%%%%%%%%%%%%%%%%%%%%%%%%%%%%%%%%%%%%%
\section{CARE}
\label{sec:method}

At each control step a CARE defender executes the same five operations upon
its own observation \eqref{eq:observation}: it updates its record of the
bearings at which it has observed arrivals and drifts its home toward the
demand that record describes, sizes an engagement region from the teammates it
detects, restricts its candidate set to the intruders inside that region,
selects from among them the cheapest interception that remains feasible, and
returns home when there is no feasible capture task available. No operation
reads a message, and
none presumes that another defender has reached a compatible conclusion.

\subsection{The engagement region}
\label{sec:method:region}

Detection establishes what exists, which is a poor guide to what a defender
should act upon. Adjacent defenders share most of their detection footprint
(Fig.~\ref{fig:setup}), so a defender that plans against everything it sees
could select the same intruder as its neighbor and leave other areas of the
perimeter unprotected. CARE therefore narrows commitment with a second and
smaller radius. The \emph{engagement radius} of defender $i$ is
the distance to its nearest detected teammate, and the detection radius when it
detects none,
\begin{equation}
\rho_i(t) =
\begin{cases}
\displaystyle \min_{k\in\mathcal{K}_i(t)} \lVert p_k(t)-p_i(t)\rVert,
& \mathcal{K}_i(t)\neq\varnothing,\\[1.0em]
r_d, & \text{otherwise,}
\end{cases}
\label{eq:engagement_radius}
\end{equation}
and the set it plans against is
\begin{equation}
\mathcal{E}_i(t) = \{\, j\in\mathcal{V}_i(t) : \lVert q_j(t)-p_i(t)\rVert \le \rho_i(t) \,\}
\subseteq \mathcal{V}_i(t).
\label{eq:engaged_set}
\end{equation}
The defender continues to sense out to the full $r_d$, since that is how it
locates the teammate of $\mathcal{K}_i(t)$ that sets $\rho_i(t)$. Because
\eqref{eq:engagement_radius} is re-evaluated from live positions at every
step, the region contracts and expands as the formation deforms. A defender
that leaves its post widens the gate of the neighbor it has moved away from,
and that neighbor can then take over the vacated arc. Freezing $\rho_i$ at
its initial value is a reasonable alternative, and is reported as an ablation
in Sec.~\ref{sec:results}.

\subsection{Target selection}
\label{sec:method:target}

For each candidate $j\in\mathcal{E}_i(t)$ the defender predicts the crossing
point and time. Under \eqref{eq:intruder_motion} the crossing point is the
radial projection of the intruder onto the perimeter and the crossing time
follows from its remaining distance,
\begin{equation}
q_j^\star(t) = \frac{R}{\lVert q_j(t)\rVert}\,q_j(t),
\qquad
\tau_j(t) = t + \frac{\lVert q_j(t)\rVert - R}{v_I}.
\label{eq:intercept}
\end{equation}
Since defenders are perimeter-constrained, the distance to a point $x$ on the
circle is the shorter arc
$\delta_i(x) = R\,|\mathrm{wrap}(\angle x - \phi_i(t))|$. Here
$\mathrm{wrap}(\cdot)$ maps an angle into $(-\pi,\pi]$ by adding or
subtracting $2\pi$, so that a difference of two bearings is always the
shorter way round the circle. A candidate is
\emph{feasible} when the defender can arrive first,
\begin{equation}
\delta_i\big(q_j^\star(t)\big)/v_D \;<\; \tau_j(t)-t ,
\label{eq:feasibility}
\end{equation}
and feasible candidates are scored by
\begin{equation}
c_i(j,t) = \Big[\delta_i\big(q_j^\star(t)\big)
+ \tfrac12 \big(\lVert q_j(t)\rVert - R\big)\Big]
\big(\tau_j(t)-t\big)^2 .
\label{eq:cost}
\end{equation}
The bracket is a distance, the defender's travel to the crossing point plus
half the distance the intruder still has to cover, so a nearer interception
scores lower. The squared time-to-crossing then shrinks that score for an
intruder about to breach, so the most imminent breach is engaged first.
Equation~\eqref{eq:cost} is not new. It is the composite loss of
DMRST-MTA~\cite{velhal2022decentralized}, the product of their spatial and
temporal loss functions, with their scaling factor set to $\tfrac12$ and
specialized to a single task: their temporal loss multiplies an intruder's
time-to-arrival by the interval a defender has after completing its previous
task, which for a first task is the whole time remaining.
Equation~\eqref{eq:feasibility} is likewise their feasibility condition.
Keeping the cost unchanged is deliberate: the comparison with DMRST-MTA in
Sec.~\ref{sec:results} then isolates a difference in coordination, not a
difference in what the two methods regard as a good interception.

A defender that re-decided from scratch at every step would oscillate
between near-tied candidates without progressing toward either, so the
previous choice $j_i^-$ receives a retention bonus $\gamma\in(0,1]$,
\begin{equation}
j_i^\star(t) = \operatorname*{arg\,min}_{j\in\mathcal{E}_i(t)\ \mathrm{feasible}}
\ \gamma^{\,\mathbb{1}[j=j_i^-]}\, c_i(j,t),
\label{eq:selection}
\end{equation}
with $j_i^\star(t)=\varnothing$ when no candidate is feasible; we use
$\gamma=0.95$. The defender then drives toward the crossing point of its target where it has
one, and toward its home at $\theta_i(t)$ otherwise, moving tangentially at
$v_D$. Returning home, instead of halting, matters: the home positions
sustain coverage between engagements, and a defender idling wherever it
happens to be leaves a gap in the perimeter.

\subsection{Homes that follow observed arrivals}
\label{sec:method:homes}

\begin{figure}[t]
\centering
\includegraphics[width=\columnwidth]{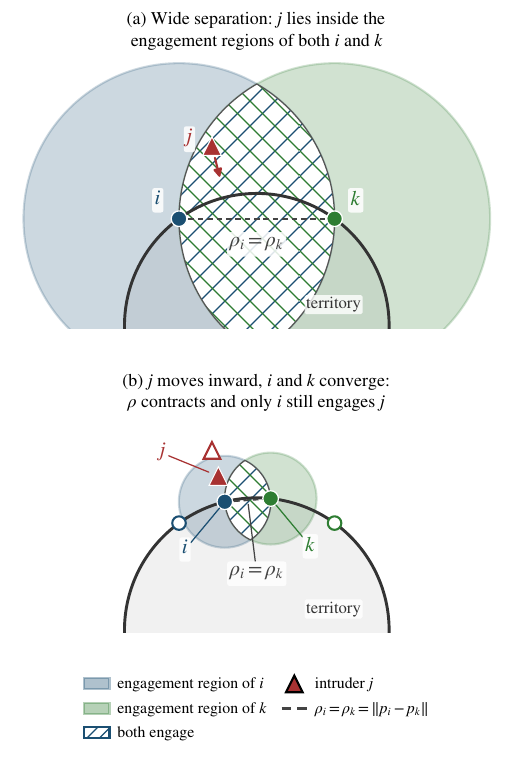}
\caption{Dissipation of a duplicate commitment. (a)~Far apart, $i$ and $k$
both hold intruder $j$ inside their engagement regions. (b)~$j$ has moved
inward and $i$ has reached its crossing point; $\rho_i=\rho_k=\lVert
p_i-p_k\rVert$ has contracted until $j$ lies outside $k$'s region, and $k$
releases it. Hollow markers: positions in~(a).}
\label{fig:dissipation}
\end{figure}

Uniformly spaced homes suit only uniform arrivals: when arrivals concentrate
in one sector, the defenders on the quiet side sit idle while the busy side
is saturated. Adaptive
partitioning~\cite{macharet2020adaptive} addresses this by moving sector
boundaries toward estimated demand, but requires the team to agree upon the
partition. We seek the same effect from what a single defender can observe for
itself.

The estimator follows~\cite{macharet2020adaptive}, which builds a
wrapped-normal density over arrival bearings from a sliding window of recent
incursions; the difference lies in whose incursions. The first time defender
$i$ detects intruder $j$ it records that intruder's bearing from the territory
center,
\begin{equation}
a_j = \operatorname{atan2}\big(q_{j,y},\ q_{j,x}\big),
\label{eq:arrival_angle}
\end{equation}
which under \eqref{eq:intruder_motion} is its spawn bearing $\psi_j$, retaining
the most recent $W$ such angles in a set $\mathcal{H}_i$, from which it forms
\begin{equation}
f_i(\alpha) \;\propto\; \sum_{a\in\mathcal{H}_i}
\exp\!\Big(-\big(\mathrm{wrap}(\alpha-a)\big)^2/2\sigma^2\Big)
\label{eq:density}
\end{equation}
on a uniform grid of $B$ bins, normalized to unit mass, with $\mathrm{wrap}$
as in Sec.~\ref{sec:method:target} so that the density is continuous across
$\pm\pi$. Nothing is shared, so
two defenders observing the same sector accumulate different records, each
biased toward its own neighborhood, which is the region its home ought to
serve.

So that a defender answers for its own stretch of perimeter, the density is
restricted to the arc bounded by the midpoints to the nearest detected teammate
on either side,
\begin{equation}
\mathcal{A}_i(t) = \big[\ \theta_i - \beta_i^-/2,\ \ \theta_i + \beta_i^+/2\ \big],
\label{eq:own_arc}
\end{equation}
where $\beta_i^{\pm}$ are the angular gaps from $\theta_i$ to those teammates.
They are located by their currently observed positions, which keeps the
mechanism within \eqref{eq:comm_free_policy}: a neighbor's home angle is
private state, unobservable whenever the neighbor is away from home. The
teammates' initial home angles are admissible, since the initial spacing is
common knowledge, and Sec.~\ref{sec:results} reports what is lost by
bounding the arc with those instead. The home drifts toward the
density-weighted centroid of the arc,
\begin{equation}
\bar\alpha_i(t) = \int_{\mathcal{A}_i(t)} \alpha\, f_i(\alpha)\,d\alpha \Big/
\int_{\mathcal{A}_i(t)} f_i(\alpha)\,d\alpha ,
\label{eq:centroid}
\end{equation}
under the rate-limited proportional law
\begin{equation}
\dot\theta_i = \operatorname{clip}\big(k_h\,\mathrm{wrap}(\bar\alpha_i(t)-\theta_i),\
-\omega_{\max},\ \omega_{\max}\big),
\label{eq:drift}
\end{equation}
with $k_h$ the proportional gain and $\omega_{\max}$ the maximum rate at
which a home is allowed to move. Equation~\eqref{eq:centroid} is a Lloyd step, as in the centroidal Voronoi
partition of~\cite{macharet2020adaptive}, but computed over one defender's own
arc and density, not over a jointly maintained partition.

The rate limit matters. If a home moves quickly, the defender resting on it
moves with it, and through \eqref{eq:engagement_radius} that changes the
engagement radii of its neighbors while they may be committed to a chase. A
gate that shifts under a committed chase is exactly the failure CARE is
meant to avoid, so $\dot\theta_i$ is bounded. The partition then stays almost
still over the length of one engagement, but still follows demand over the
length of an episode.

The two constants have units. A gain in $\mathrm{s^{-1}}$ and a cap in
$\mathrm{rad\,s^{-1}}$ would have to be re-tuned whenever $R$ or $v_I$
changed, so we express both as multiples of $v_D/R$, the fastest a defender
can move around the perimeter in angle:
\begin{equation}
k_h = 5.0\,v_D/R, \qquad \omega_{\max} = 0.5\,v_D/R .
\label{eq:drift_gains}
\end{equation}
The simulation uses the fixed values $k_h=0.5\,\mathrm{s^{-1}}$ and
$\omega_{\max}=0.05\,\mathrm{rad\,s^{-1}}$, which are what
\eqref{eq:drift_gains} gives at the benchmark's nominal $v_D=v_I$. In the
scenarios that vary $v_D$ the drift is therefore not rescaled with the
defender, whereas adaptive partitioning does rescale its own rate limit with
$v_D$. The hardware demonstration of Sec.~\ref{sec:hardware} uses the form of
\eqref{eq:drift_gains}, which is what makes the transfer to a $0.5$~m
perimeter possible. The remaining constants are $\sigma=0.5$~rad, $W=25$ and
$B=180$.

\subsection{Algorithm and cost}
\label{sec:method:algo}

\begin{algorithm}[t]
\caption{CARE: one control step at defender $i$}
\label{alg:step}
\begin{algorithmic}[1]
\REQUIRE own state $p_i,\theta_i$; observation $o_i(t)$; previous target $j_i^-$
\FOR{$j \in \mathcal{V}_i(t)$ not previously detected}
  \STATE append $a_j$ of \eqref{eq:arrival_angle} to $\mathcal{H}_i$, keeping the last $W$
\ENDFOR
\IF{$\mathcal{H}_i \neq \varnothing$ \AND $\mathcal{K}_i(t) \neq \varnothing$}
  \STATE drift the home: $\theta_i \leftarrow \theta_i + \dot\theta_i\,\Delta t$ by \eqref{eq:density}--\eqref{eq:drift}
\ENDIF
\STATE $\rho_i \leftarrow$ distance to nearest detected teammate, else $r_d$, by \eqref{eq:engagement_radius}
\STATE $\mathcal{E}_i \leftarrow \{ j\in\mathcal{V}_i(t) : \lVert q_j-p_i\rVert \le \rho_i \}$
\STATE $c^\star \leftarrow \infty$; \ $j^\star \leftarrow \varnothing$
\FOR{$j \in \mathcal{E}_i$}
  \STATE $(q_j^\star,\tau_j) \leftarrow$ predicted crossing of $j$, by \eqref{eq:intercept}
  \IF{$\delta_i(q_j^\star)/v_D < \tau_j - t$}
    \STATE $c \leftarrow c_i(j,t)$, discounted by $\gamma$ if $j = j_i^-$
    \STATE \textbf{if} $c < c^\star$ \textbf{then} $c^\star \leftarrow c$; \ $j^\star \leftarrow j$
  \ENDIF
\ENDFOR
\RETURN tangential velocity at $v_D$ toward $q^\star_{j^\star}$ if
$j^\star \neq \varnothing$, else toward the home at $\theta_i$
\end{algorithmic}
\end{algorithm}

Algorithm~\ref{alg:step} is one control step for one defender. Every
defender runs it independently, and in parallel. Its cost is found as follows.
Sizing $\rho_i$ scans at most $N_d-1$ teammates. Forming $\mathcal{E}_i$
scans at most $V$ intruders, where $V=\max_i\lvert\mathcal{V}_i(t)\rvert$
is the number currently visible. Scoring is a single pass over
$\mathcal{E}_i$, and the density costs $O(WB)$ with $W$ and $B$ fixed. A
team-wide step is therefore $O(N_d(N_d+V+WB))$, and nothing in it iterates
to convergence. A consensus-based auction has no bound of this kind. Each
bundle layer re-scores an insertion for every candidate, and the
reconciliation that follows sweeps every pair of defenders over every bid,
for as many rounds as it takes to agree; that number is not known in
advance.

\subsection{Dissipation of duplicate commitments}
\label{sec:method:conflicts}

Two defenders can select the same intruder. Their engagement regions
overlap and neither sees the other's choice, so nothing above prevents it.
What the construction does provide is decay. Suppose defenders $i$ and $k$
both commit to intruder $j$, and $i$ is the closer
(Fig.~\ref{fig:dissipation}). Both move toward the same crossing point, so
the distance between them falls, and by \eqref{eq:engagement_radius} both
engagement radii shrink with it. Because $j$ is farther from $k$ than from
$i$, $k$ loses $j$ from $\mathcal{E}_k$ first. It then re-solves
\eqref{eq:selection} over what remains and either picks a different intruder
or returns home. The duplicate is resolved because the defenders moved, not
because either was told anything. This is a weaker guarantee than an auction
gives. Sec.~\ref{sec:results} shows that duplicates are short-lived and
cheap: the communication that would prevent them outright is worth $0.0004$
in capture rate.

Hence the name. The planning is \emph{communication-free} because no step of
Algorithm~\ref{alg:step} reads a message. The \emph{regions of engagement}
are \emph{adaptive} because each is re-sized from the live geometry at every
step, and it is that re-sizing, not any exchange between defenders, that
dissipates the duplicates.

%%%%%%%%%%%%%%%%%%%%%%%%%%%%%%%%%%%%%%%%%%%%%%%%%%%%%%%%%%%%%%%%%%%%%%%%%%%%%%%%
\section{Results}
\label{sec:results}

\subsection{Setup}
\label{sec:results:setup}

All simulations ran on a dual-socket Intel Xeon Gold 5320 workstation
(52 cores, 256~GiB), with episodes distributed over 100 worker processes, so
the per-step timings in Sec.~\ref{sec:results:main} are measured under
contention.

The benchmark comprises
193 scenarios, each run for 200 episodes, which gives 38{,}600 episodes per
method. Episodes are \emph{seed-paired}: for a given scenario and episode index every
method sees an identical world, down to the defenders' initial angles and each
intruder's spawn position, time and speed; a fingerprint taken at every reset
verifies this. All differences below are paired and per-episode.

The scenarios sweep the dimensionless groups of \eqref{eq:groups}: team size
$N_d\in\{3,4,5,6\}$, load $\lambda R/v_I$ from $1.0$ to $13.3$, and speed ratio
$v_D/v_I\in\{0.8,1.0,1.25,1.5,2.0\}$. Arrivals are uniform in 165 scenarios and
von Mises in 28, with concentration $\kappa\in\{1,2,4,6,8\}$. Ten scenarios replace the radial intruder of \eqref{eq:intruder_motion} with a
zigzagging or spiralling one, so that every method's crossing prediction is
wrong. Throughout, $r_d/R=4$, $r_c/R=0.04$, $R_s=6R$, and the horizon is
$60\,R/v_I$.

We report the capture rate $\eta$ of \eqref{eq:capture_rate}. Every comparison is stated as $\Delta\eta = \eta_{\mathrm{CARE}} -
\eta_{\mathrm{other}}$, taken within each shared episode and then averaged
over them. \textbf{A positive $\Delta\eta$ is a win
for CARE, meaning it captured a larger fraction of intruders; a negative
$\Delta\eta$ is a loss.} A per-scenario $\Delta\eta$ is significant when its
95\% confidence interval excludes zero,
and the w/l column of every table below counts scenarios on which CARE is
significantly better and significantly worse.

One caveat. DMRST-MTA is stated for defenders faster than intruders,
$v_D>v_I$~\cite{velhal2022decentralized}, and 20 of our 193 scenarios set
$v_D/v_I=0.8$. We keep them, since a defense that works only when it is
faster than the threat is of limited use, but on those scenarios the baseline
runs outside its published operating range. Adaptive partitioning is not
affected: its reachability test is written for an arbitrary speed
ratio~\cite{macharet2020adaptive}.

DMRST-MTA with full communication exchanges bids among all defenders within
$r_d$, which on a perimeter is the whole team, and iterates its consensus
until the bids agree or a cap of 100 rounds per bundle layer is reached. The
cap is ours; the original paper states none and reports no timing. Where we
logged it, the cap was reached on a minority of layers. It also sets that method's per-step cost, so Table~\ref{tab:main} marks that
entry and we draw no comparison from it.

\begin{table*}[!t]
\centering
\caption{Capture rate over the 193 constant-velocity scenarios, 38{,}600
seed-paired episodes per comparison. $\Delta\eta$ and w/l are as defined in
Sec.~\ref{sec:results:setup}: positive $\Delta\eta$ favors CARE.
$^\dagger$Measured under the consensus cap of Sec.~\ref{sec:results:setup},
which is ours. The one-hop rows were timed in a separate run and are not
compared on computational cost.}
\label{tab:main}
\begin{tabular}{lrrrrr}
\toprule
Method & $\eta$ & $\Delta\eta$ & 95\% CI & w/l & ms/step \\
\midrule
DMRST-MTA, no communication & 0.622 & $+0.1654$ & $[+0.1640, +0.1668]$ & 193/0 & 80.3 \\
DMRST-MTA, one hop & 0.740 & $+0.0472$ & $[+0.0463, +0.0481]$ & 184/1 & -- \\
DMRST-MTA, full communication & 0.782 & $+0.0048$ & $[+0.0039, +0.0058]$ & 54/76 & 439.0$^\dagger$ \\
Adaptive partitioning & 0.775 & $+0.0117$ & $[+0.0110, +0.0123]$ & 116/38 & 6.3 \\
\midrule
\textbf{CARE} & \textbf{0.787} & & & & \textbf{4.2} \\
\midrule
CARE without home drift & 0.739 & $+0.0485$ & $[+0.0477, +0.0494]$ & 190/0 & 2.8 \\
CARE, arcs from initial homes & 0.763 & $+0.0242$ & $[+0.0236, +0.0248]$ & 178/0 & 4.1 \\
CARE, frozen engagement radius & 0.749 & $+0.0383$ & $[+0.0375, +0.0390]$ & 171/0 & 4.2 \\
CARE $+$ one hop & 0.787 & $-0.0004$ & $[-0.0005, -0.0003]$ & 1/19 & -- \\
\bottomrule
\end{tabular}

\end{table*}

\subsection{Comparison with the baselines}
\label{sec:results:main}

Table~\ref{tab:main} gives the headline comparison. Against DMRST-MTA with its messages removed, CARE is ahead by $0.165$ and
significantly better on every one of the 193 scenarios, with no losses
anywhere in the benchmark. The narrowest margin, on the three-defender
scenario at the highest arrival rate, is $+0.012$ and is itself significant.
This is the like-for-like comparison: two decentralized policies, neither
permitted to communicate.

The gap closes only when the auction is permitted to talk without
restriction. One hop of communication recovers much of DMRST-MTA's loss, from
$0.622$ to $0.740$, and full consensus a further $0.042$, to $0.782$, which
leaves CARE marginally ahead. The mean difference of $0.005$ in CARE's favor
is significant, but it hides a split: the auction is significantly ahead on 76
scenarios and behind on 54, and which side wins depends on the arrival
load. On the 39 scenarios at $\lambda R/v_I=1$
CARE leads by $0.085$ and on every one of them, whereas at each higher load
the auction leads by $0.010$ to $0.023$. In other words, the auction needs
every message delivered to match a method that sends none, and what then
separates the two is a trade-off across load, not a gap.

Adaptive partitioning is the harder comparison, and the margin is narrow.
Running with its coordination intact it reaches $0.775$ against CARE's
$0.787$, a mean difference of $0.012$. That average conceals structure, disaggregated in
Sec.~\ref{sec:results:teamsize}: CARE is significantly better on 116
scenarios and worse on 38, the worst being $-0.035$.

The lower half of Table~\ref{tab:main} removes one component at a time.
Removing the home drift and holding homes at their initial angles costs
$0.049$, the largest single effect. Bounding each defender's arc with the
teammates' initial home angles instead of their live observed positions costs
$0.024$: those angles are common knowledge, but once defenders have moved they
no longer describe where anyone is. Freezing the engagement radius at its initial value costs
$0.038$.

CARE is also the cheaper policy to run, as the last column of
Table~\ref{tab:main} records: adaptive partitioning costs $1.5$ times as much
compute time per control step and DMRST-MTA $19$ times as much, the ratio the
complexity argument of Sec.~\ref{sec:method:algo} predicts. All three are
research implementations timed under the same conditions, so the ratios, not
the absolute figures, are the meaningful result. The auction with full
communication is not compared on cost, since its figure is set by our
consensus cap.

\begin{table}[!t]
\centering
\caption{CARE against adaptive partitioning by team size. The right-hand
columns restrict to the family holding load fixed at four intruders per
defender, separating team size from load.}
\label{tab:teamsize}
\begin{tabular}{crrrrr}
\toprule
& \multicolumn{3}{c}{all scenarios} & \multicolumn{2}{c}{fixed load} \\
\cmidrule(lr){2-4}\cmidrule(lr){5-6}
$N_d$ & $n$ & $\Delta\eta$ & w/l & $\Delta\eta$ & w/l \\
\midrule
3 & 40 & $-0.0134$ & 5/26 & $-0.0132$ & 1/6 \\
4 & 35 & $+0.0031$ & 13/9 & $+0.0057$ & 3/1 \\
5 & 83 & $+0.0212$ & 66/3 & $+0.0154$ & 7/1 \\
6 & 35 & $+0.0263$ & 32/0 & $+0.0265$ & 10/0 \\
\bottomrule
\end{tabular}

\end{table}

\subsection{What a hop of communication would add}
\label{sec:results:onehop}

A comparison across methods confounds communication with everything else
that differs between them, so the claim that CARE does not need
communication is best tested against CARE itself. We therefore ran an
ablation identical in every respect but one: each defender broadcasts the
identity of its current target to teammates within $r_d$, and drops a
candidate already claimed by a closer one, the duplicate commitments of
Sec.~\ref{sec:method:conflicts}.

Communication does help, and the size of the help is the finding. CARE trails
the ablation by $0.0004$ over 38{,}600 paired episodes, with a 95\% interval
of $[-0.0005,-0.0003]$, and is significantly worse on 19 of the 193 scenarios
and significantly better on one. Set that against the $0.165$ gap to DMRST-MTA without messages, and against the
$0.161$ that restoring its messages is worth to that auction on the same
episodes: the coordination CARE forgoes is not coordination it needs.

\begin{table}[!t]
\centering
\caption{Gain from the home-drift mechanism alone, on the 5v25 scenarios with
$\mu=0$. Entries are $\Delta\eta$ between CARE and CARE with homes held fixed;
$\kappa$ is the von Mises concentration, higher being a narrower sector of
attack.}
\label{tab:vonmises}
\begin{tabular}{crrr}
\toprule
& \multicolumn{3}{c}{$\lambda R / v_I$} \\
\cmidrule(lr){2-4}
$\kappa$ & 1.00 & 6.67 & 13.33 \\
\midrule
1 & $+0.0398$ & $+0.0714$ & $+0.0672$ \\
2 & $+0.0684$ & $+0.1372$ & $+0.1612$ \\
4 & $+0.1124$ & $+0.1948$ & $+0.1974$ \\
6 & $+0.1322$ & $+0.2380$ & $+0.2094$ \\
8 & $+0.1346$ & $+0.2560$ & $+0.2464$ \\
\bottomrule
\end{tabular}

\end{table}

\subsection{Where the partitioning comparison turns}
\label{sec:results:teamsize}

The mean advantage over adaptive partitioning is small, and it is not uniform
across the benchmark. Table~\ref{tab:teamsize} splits it by team size. At six defenders CARE leads
by $0.0263$ and is significantly better on 32 of 35 scenarios with no losses;
at five it leads by $0.0212$ and is better on 66 of 83; at four the two
methods are indistinguishable on average; and at three defenders CARE is
\emph{worse}, by $0.0134$, losing 26 of 40 scenarios.

Team size and load per defender are confounded in that breakdown, since
fewer defenders generally means more intruders each. To separate them, the
right-hand columns of Table~\ref{tab:teamsize} restrict to the family holding
the ratio at four intruders per defender: the gradient survives almost
unchanged, $-0.0132$ at three defenders rising to $+0.0265$ at six. The
effect is team size, not load.

A plausible explanation lies in how the arcs of \eqref{eq:own_arc} are
built. Each defender's arc is bounded by one teammate on either side. With
three defenders every arc spans $2\pi/3$ and every defender is a boundary for
both of the others, so one defender drawn far from home distorts the whole
team's partition at once. Adaptive partitioning maintains its partition by
agreement and does not suffer from this.

\subsection{Concentrated arrivals}
\label{sec:results:vonmises}

Uniformly spaced homes handle uniform arrivals well and anything else
badly, and that is the case the home drift is designed for. All 28 von Mises
scenarios are 5v25, so this section speaks to one team size. Across them,
CARE leads DMRST-MTA without communication by $0.200$ and adaptive
partitioning by $0.022$; it is better on all 28 and on 25 of 28
respectively, with no losses in either. The second result is the one worth
noting. Redistributing defenders toward demand is what adaptive partitioning
was built to do, and it does so by agreement between the defenders, whereas
CARE does so from each defender's own observations. The auction with full
communication is the one method ahead of CARE here, by $0.036$ across the 28
scenarios and by $0.064$ at $\kappa=8$. That margin grows with concentration
and, at $\kappa=8$, with load as well. The drift recovers most of what
coordination provides under concentrated arrivals, but not all of it.

Table~\ref{tab:vonmises} shows what the home drift alone is worth. The
gain grows with concentration at every
arrival rate, from $+0.040$ when arrivals are spread widest to $+0.256$ when
they are narrowest. It also grows with load, though at $\kappa=8$ not in
every step. Both trends are what one would expect. The narrower the sector
of attack, the worse fixed uniform homes are placed; and the higher the
arrival rate, the less time there is to make up for being in the wrong
place.

\subsection{Maneuvering intruders}
\label{sec:results:evasive}

Every method considered here predicts a crossing point by extrapolating the
radial motion of \eqref{eq:intruder_motion}, so the ten scenarios whose
intruders zigzag or spiral inward along routes fixed at spawn test what
happens when that prediction is wrong. All three methods lose capture rate
relative to the straight-line scenarios, but the losses are not equal: adaptive partitioning falls by $0.056$ and
DMRST-MTA without communication by $0.043$, whereas CARE falls by $0.021$.
The margin over adaptive partitioning consequently widens from $0.012$ across
the benchmark as a whole to $0.045$ here, significant on all ten scenarios.

The asymmetry follows from how each method decides ownership. A partition
assigns an intruder to a defender by the sector into which its bearing falls,
so an intruder that changes bearing in flight can migrate between sectors
while being pursued. The engagement region of \eqref{eq:engagement_radius}
is a distance test, indifferent to bearing: a defender's claim on an intruder
is re-evaluated against the current geometry at every step, so it follows the
intruder as it maneuvers instead of resting on a sector assignment the
maneuver has invalidated. The arrival-density estimator is likewise
unaffected, since \eqref{eq:arrival_angle} records a bearing once, at first
detection.

\subsection{Hardware demonstration}
\label{sec:hardware}

\begin{figure}[t]
\centering
\includegraphics[width=0.99\columnwidth]{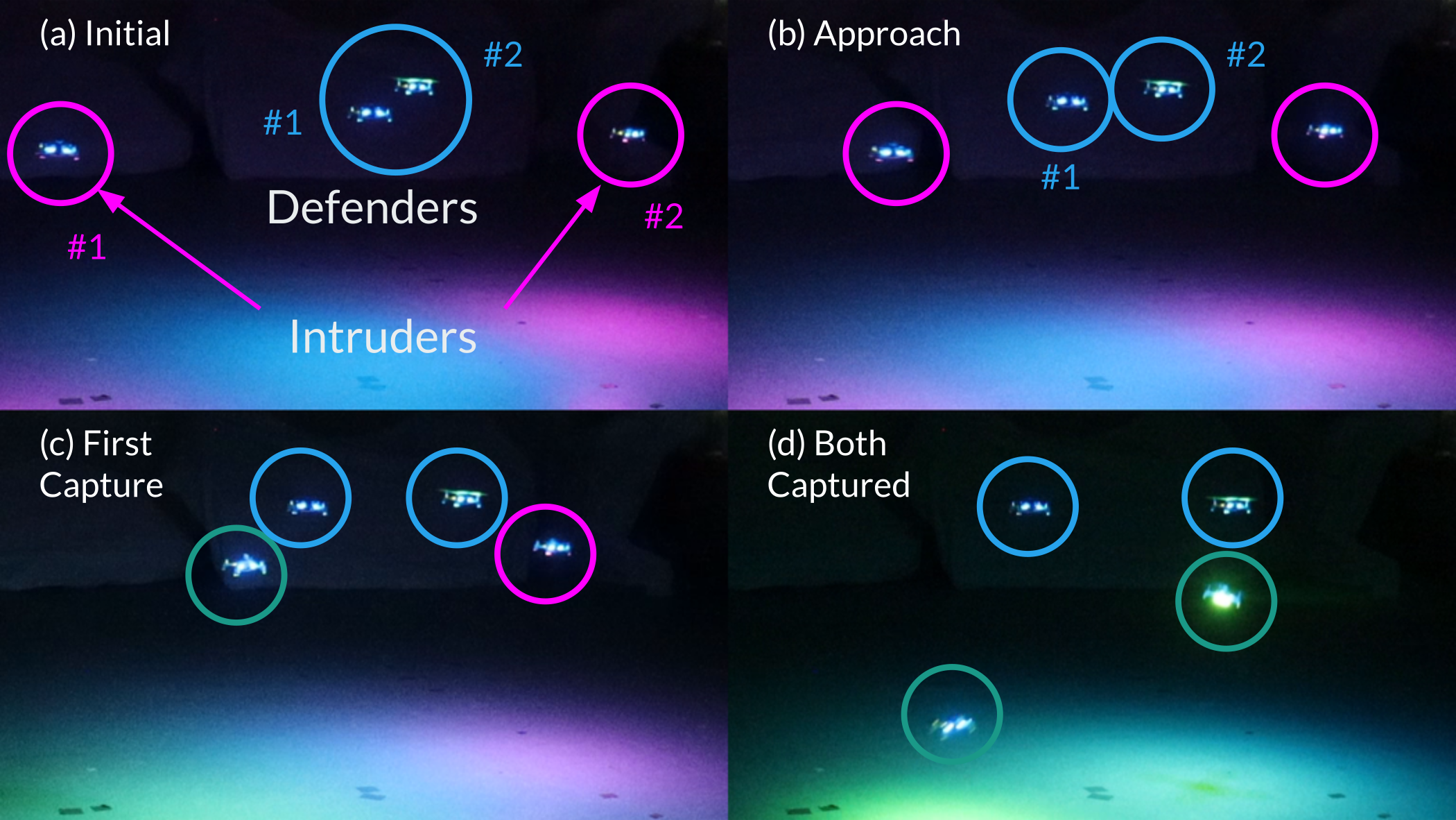}
\caption{\textbf{CARE hardware demonstration with a Crazyflie team.}
Two blue defenders maintain a $0.5\,\mathrm{m}$ protected perimeter against two
magenta intruders following scripted inbound trajectories. The sequence shows
(a) the initial configuration, (b) the approach, (c) the first capture as the
second defender intercepts the remaining intruder, and (d) both intruders
captured. Upon capture, intruders turn green and descend to avoid unintended collisions.}
\label{fig:hardware}
\end{figure}

CARE is demonstrated on hardware as a feasibility check. The flight code
is a separate implementation of Algorithm~\ref{alg:step} for the Crazyswarm2
stack. It shares the simulation's specification and parameters. The
platform is a team of Crazyflie quadrotors
under Qualisys motion capture, every agent a vehicle, intruders included
(Fig.~\ref{fig:hardware}), so team size is bounded by the airframes
available: two defenders against two intruders. The protected radius is
$R=0.5$~m, set by the workspace, and the arrival rate $\lambda=0.2$. The
parameters are carried over through the
dimensionless groups of \eqref{eq:groups} rather than retuned, with the drift
gains transferred in the form of \eqref{eq:drift_gains}.

%%%%%%%%%%%%%%%%%%%%%%%%%%%%%%%%%%%%%%%%%%%%%%%%%%%%%%%%%%%%%%%%%%%%%%%%%%%%%%%%
\section{Conclusion}
\label{sec:conclusion}

CARE shows that a team of defenders can divide a perimeter between them
without exchanging messages. Each defender senses widely but commits
only within a region sized by the distance to its nearest teammate, and lets
its resting position follow the arrivals it has seen for itself. Conflicts
are not negotiated away; they dissolve as the defenders move. In simulation
CARE, without any communication, performs on par with or better than an
adaptive partitioning method and a bundle auction, both run with their
communication intact. Against the same auction with its messages removed,
CARE is ahead by $0.165$ and wins on all 193 scenarios. The sharpest test is
the ablation: adding one hop of communication raises CARE's capture rate by
$0.0004$, whereas removing the auction's messages lowers its own by $0.161$,
which shows that CARE is, by design, communication-agnostic. There are however some limitations: the agents are
first-order, so vehicles with a significant acceleration time fall outside
the setting. And with $r_d=4R$ every defender sees every teammate perfectly;
noisy or intermittent detection is untested.

\bibliographystyle{ieeetr}
\bibliography{references}

\end{document}